\documentclass[10pt,twocolumn,letterpaper]{article}

\usepackage[pagenumbers]{cvpr} 

\usepackage{graphicx}
\usepackage{amsmath}
\usepackage{amssymb}
\usepackage{booktabs}
\usepackage[accsupp]{axessibility}

\usepackage[longend]{algorithm2e}
\RestyleAlgo{ruled}

\usepackage[pagebackref,breaklinks,colorlinks]{hyperref}

\usepackage[capitalize]{cleveref}
\crefname{section}{Sec.}{Secs.}
\Crefname{section}{Section}{Sections}
\Crefname{table}{Table}{Tables}
\crefname{table}{Tab.}{Tabs.}

\def\cvprPaperID{6486} 
\def\confName{CVPR}
\def\confYear{2022}

\title{SkinningNet: Two-Stream Graph Convolutional Neural Network for\\Skinning Prediction of Synthetic Characters}

\begin{document}


\author{Albert Mosella-Montoro and Javier Ruiz-Hidalgo\\
Universitat Polit\`ecnica de Catalunya\\
{\tt\small \{albert.mosella, j.ruiz\}@upc.edu}\\
{\tt\small \url{https://imatge-upc.github.io/skinningnet}} \\}


\maketitle

\begin{abstract}
This work presents SkinningNet, an end-to-end Two-Stream Graph Neural Network architecture that computes skinning weights from an input mesh and its associated skeleton, without making any assumptions on shape class and structure of the provided mesh. Whereas previous methods pre-compute handcrafted features that relate the mesh and the skeleton or assume a fixed topology of the skeleton, the proposed method extracts this information in an end-to-end learnable fashion by jointly learning the best relationship between mesh vertices and skeleton joints. The proposed method exploits the benefits of the novel Multi-Aggregator Graph Convolution that combines the results of different aggregators during the summarizing step of the Message-Passing scheme, helping the operation to generalize for unseen topologies. Experimental results demonstrate the effectiveness of the contributions of our novel architecture, with SkinningNet outperforming current state-of-the-art alternatives.
\end{abstract}
\section{Introduction} \label{sec:intro}

Animating a 3D character is a complex and time-consuming process that animators spend years learning to do efficiently. In a typical animation pipeline, an artist first creates a mesh model and specifies the skeleton topology, skinning weights are then painted manually. During this process, the animators follow two main steps. First, they do the skin binding, which consists of defining which parts of the mesh will be affected by the movement of a specific joint. Finally, they decide the skinning weights that describe which amount of movement is transferred to the skin.
\begin{figure}[ht]
 \centering
 \includegraphics[width=0.45\textwidth]{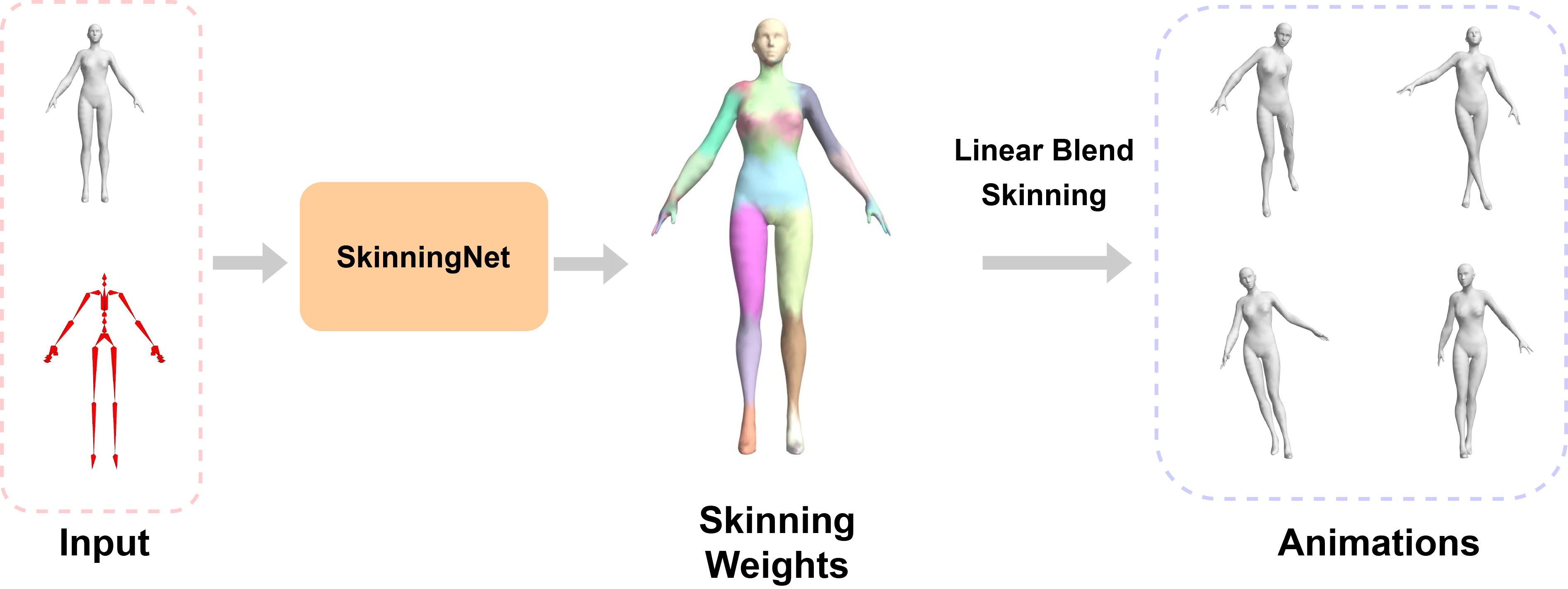}
 \caption{\textbf{Overview} of the proposed pipeline. Given an input mesh and its associated skeleton, SkinningNet predicts the skinning weights used by the Linear Blend Skinning~\cite{lbs} algorithm to create a set of animations.}
 \label{fig:teaser}
\end{figure}

SkinningNet is a Two-Stream Graph Convolutional Network that, given an input mesh and its corresponding skeleton, performs the skin binding and then computes the skinning weights for each mesh vertex as it can be seen in Fig.~\ref{fig:teaser}. Previous methods rely on pre-computed handcrafted features to create the mesh and skeleton relationship. The proposed method automatically extracts the best features to relate the mesh and the skeleton to predict the associated skinning weights. The main contributions of this paper are: a) A Two-Stream Graph Neural architecture that learns to extract features from meshes and skeletons with different topologies, b) A Multi-aggregator Graph Convolution (MAGC) layer that extends the Message-Passing scheme to use a multiple aggregation approach that better generalizes for unseen graph topologies, c) A novel skin binding method that uses a skeleton joint representation instead of a bone representation and d) A Mesh-Skeleton Graph Convolution Network that exploits this skeleton joint representation to find the optimal relationship between the input mesh and the skeleton. All these contributions allow the proposed method to outperform current state-of-the-art with over $20\%$ improvement on mesh deformation error. Furthermore, experimental results show the ability of the proposed method to generalize for complex mesh and skeleton structures of different domains.
\section{Related Work}
\label{sec:sota}
This section reviews techniques related to Graph Neural Network solutions that are used for Geometric Learning and then transitions to more specific techniques used for skinning weight prediction.

\subsection{Graph Neural Networks}
Graph Neural Networks can be divided into two sub-sets: the ones that generalize the convolution operation using a spectral approach~\cite{BrunaSpectralNetworks, DefferrardCNNGSpectral} and the ones that use a spatial approach~\cite{DuvenaudCNG, kipf2017semi, MoNet}. The vast majority of the new proposed graph convolutions can be seen as an adaptation of the Message-Passing scheme~\cite{messagepassing}. The Graph Convolution Network (GCN)~\cite{kipf2017semi} is one of the popular message-passing implementations that proposes to transform the nodes of a neighbourhood using a weight matrix that is normalized by the degree of the neighbourhood. The transformed node features are aggregated using the addition operator. 

Another widely adopted graph convolution is Graph Attention Network (GAT)~\cite{GAT2018} that proposes learning an attention coefficient using edge attributes, which are defined as the difference between the central node and its neighbours. Edge Convolution Network~\cite{dgcnn} is one of the implementations widely adopted on 3D Geometric tasks, which consists of defining the edge attributes of a graph using an asymmetric function. Edge attributes of each neighbourhood are fed into a shared Multi-Layer Perceptron (MLP) and aggregated using maximum or addition aggregators. Residual connections~\cite{ragc, li2019deepgcns, Dang_2021_ICCV} have also been shown to help achieve better performance on deep Graph Neural Networks. Recently, Multi-Neighbourhood Graph Convolutions~\cite{MOSELLAMONTORO2021} introduced the combination of multiple neighbourhoods to create an enriched node descriptor. 

Most of the previously described works use mean, maximum and addition aggregators, which fail to distinguish between neighbourhoods with identical features but different cardinalities, as proved by Xu~\etal~\cite{xu2018gin}. To solve that, Dehmamy~\etal~\cite{dehmamy2019understanding} proposed to use multiple aggregators. We propose an extension of the multi-aggregator scheme using complementary aggregators and learning how to combine them instead of just concatenating or adding the results of each of them. The motivation for the use of multiple scalers is to improve the generalization of the convolution for unseen topologies while avoiding the value of each aggregation exploding when the degree of the neighbourhood increases. In SkinningNet, this extension allows the network to generalize for more complex and unseen mesh and skeleton topologies.

\subsection{Skinning Weight Prediction}

The techniques used to compute the skin deformation of synthetic characters are usually needed by real-time applications, such as video-games. Approaches such as Linear Blend Skinning (LBS)~\cite{lbs} or Dual Quaternion Skinning (DQS)~\cite{dqs} are widely used due to their simplicity and computational efficiency. These techniques compute the deformation of the mesh based on a set of skinning weights that are assigned to each of the mesh vertices. The skinning weights are either painted manually by an animator or automatically generated. Automatic skinning weight prediction techniques can be grouped into two different categories: geometric based methods~\cite{dqs,Kavan12,Wareham08,Jacobson11,Bang18,Dionne14,Mukai16,Si15,Komaritzan2018,Komaritzan2019FastPS} and data-driven solutions~\cite{Loper15,Doug05,Le14seq, Kim17datadriven, qiao2018learning}.

\noindent{\textbf{Geometric based methods}} rely on geometric characteristics between meshes and skeletons. The earliest methods to automatically generate skinning weights proposed to exploit Heat Diffusion\cite{heatdifussion} and Illumination~\cite{illuminationmodel} models. Alternatively, other methods used energy functions for the estimation, such as Elastic Energy~\cite{Kavan12} or Laplacian Energy\cite{Jacobson11}. Later, Dionne~\etal~\cite{Dionne14} proposed Geodesic Voxel Binding to handle non-watertight meshes. All these methods rely on functions that assign the skinning weights depending on the distance between joints and vertices. However, this assumption does not work in practice for AAA game characters that commonly have complex topologies where multiple independent components can intersect.

\begin{figure*}[ht]
 \centering
 \includegraphics[width=1.0\textwidth]{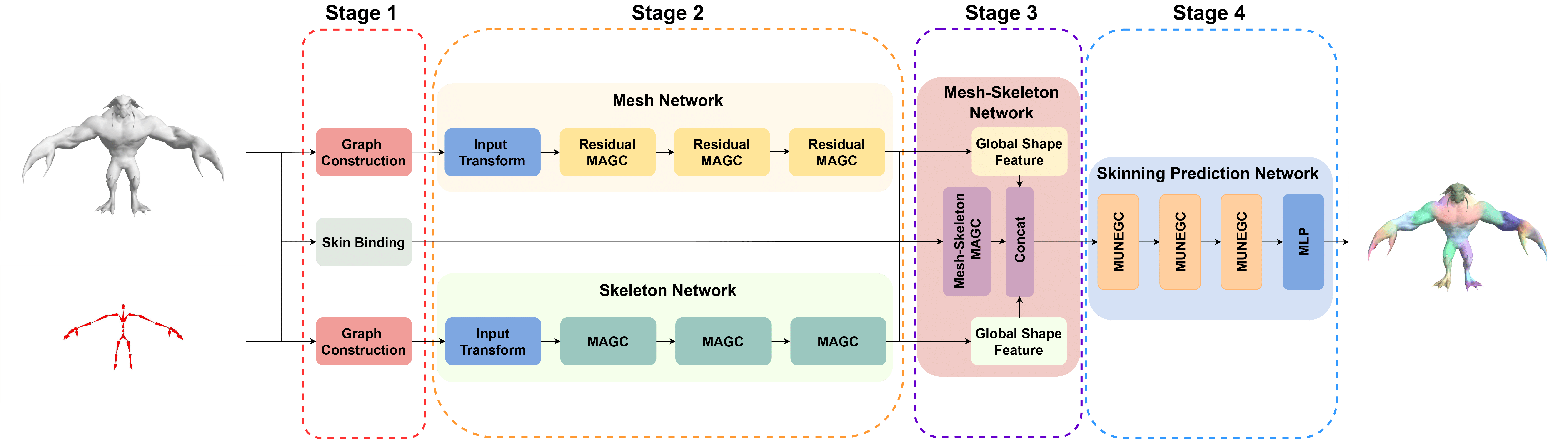}
 \caption{\textbf{SkinningNet} architecture is composed of four main stages. Stage 1 is in charge of building the needed graphs from the input mesh and its associated skeleton. Stage 2 is responsible for extracting features independently for the mesh and skeleton. Stage 3 combines the previous mesh and skeleton features to extract a descriptor that relates both structures. Stage 4 predicts the skinning weights.}
 \label{fig:architecture}
\end{figure*}

\noindent{\textbf{Data-driven methods}} typically require multiple poses of a mesh or different meshes as input to learn how to compute the skinning weights. New methods such as~\cite{li2021learning, chen2021snarf, Saito:CVPR:2021} estimate skinning weights from Motion Capture data. They focus on finding skinning weights of humanoids and assume a fixed skeleton topology, making the network unable to generalize for characters with different skeletons. 

NeuroSkinning~\cite{Liu2019} is one of the earliest proposed methods to automatically compute the skinning weights for synthetic characters using neural networks. This method makes use of graph convolutions to compute the skinning weights of a new mesh. NeuroSkinning uses a bone representation, meaning that the network predicts the skinning weights per bone instead of per joint. It also relies on creating a super-skeleton that consists of the fusion of all the skeletons that can be found in the training set. This super-skeleton is needed to cope with the fixed output of the network. As a result, this assumption makes the network unsuitable for working with skeletons that can not fit in the super-skeleton structure. 

RigNet~\cite{RigNet} tries to overcome this limitation using a k-NN approach. The network predicts the skinning weights only for the k-nearest bones to the mesh vertex. This feature allows the network to work with unseen skeleton topologies, however, the bone representation is still used. By definition, the movement of a skeleton driven mesh comes from the rotation of a joint. In an ideal scenario where each of the joints has one associated bone, both representations are equivalent. However, this assumption does not work for complex meshes, where a joint has more than one associated bone. The bone representation used in RigNet has problems to manage these scenarios, common in stylized characters. To overcome this problem, our proposal directly uses a skeleton joint representation that can manage complex skeleton topologies where each of the joints can have more than one bone. Furthermore, both RigNet and Neuroskinning rely on handcrafted features to learn the relation between a mesh and its associated skeleton. Our proposal consists of a Two-Stream Graph Neural Network that learns the relation between mesh and skeleton on training time, selecting the best set of features automatically without having to rely on selected handcrafted features.
\section{SkinningNet}

This section details the proposed SkinningNet architecture. Sec.~\ref{subsec:architecture} gives a general overview of the four stages of the proposed architecture. Sec.~\ref{subsec:graph_construction} explains how neighbourhoods are selected in the graph's creation step from the input meshes and their associated skeletons. Sec.~\ref{subsec:magc} formulates a novel graph convolution layer, \emph{Multi-Aggregator Graph Convolution} (MAGC), where multiple aggregators are used to allow the network to distinguish between neighbourhoods with similar features but different cardinality. Finally, Sec.~\ref{subsec:graph_blocks} explains how the different Graph Convolutional Blocks in SkinningNet are constructed based on the proposed MAGC.

\subsection{Architecture Overview} \label{subsec:architecture}

SkinningNet is a Two-Stream Graph Convolutional Neural Network, that takes as input a mesh and its corresponding skeleton and predicts a set of skinning weights, one for each mesh vertex. It is composed of four different stages as depicted in Fig.~\ref{fig:architecture}.

\noindent\textbf{Stage 1:} \emph{Graph Construction} and the \emph{Skin Binding} (explained in detail in Sec.~\ref{subsec:graph_construction}). The Graph Construction step converts the mesh and the skeleton inputs into two independent graphs. The Skin Binding step decides which joints will influence each vertex and creates a graph representing this relationship. The Graph Construction output is fed into Stage 2 and the output of Skin Binding is used in Stage 3.

\noindent\textbf{Stage 2:} \emph{Mesh and Skeleton Networks}. These networks transform the node attributes to feature vectors through an input transform implemented using an MLP. Each network is responsible for extracting features independently for the mesh and skeleton. The \emph{Mesh Network} is composed of three Residual MAGC layers, whereas the \emph{Skeleton Network} is composed of three MAGC layers, further details are given in Sec.~\ref{subsec:magc} and~\ref{subsec:graph_blocks}. This difference is mainly because the skeleton is usually much simpler than the mesh and does not require a deep network to learn the characteristics of its geometry. The output of both networks will be combined in the Stage 3.

\noindent\textbf{Stage 3:} \emph{Mesh-Skeleton Network}, based on a single Mesh-Skeleton MAGC layer. This block relates the mesh and the skeleton using the output of the \emph{Skin Binding} step. The output of this block is a single graph where each node can represent the vertices of the mesh or the joints of the skeleton. However, only the nodes representing the vertices of the mesh are used on the following stages. Furthermore, to help with the final skinning weight prediction in Stage 4, a global shape descriptor that encodes the global information of the mesh and skeleton graphs is extracted and concatenated to the mesh nodes. 

\noindent\textbf{Stage 4:} \emph{Skinning Prediction Network}. The final vertex skinning weight is predicted for each vertex of the mesh. It is composed of three Multi-Neighbourhood Graph Convolution (MUNEGC) layers followed by an MLP. Here, the MAGC is exploited in a multi-neighbourhood fashion, combining the mesh topology and the local shape information to extract an enriched descriptor used to predict the skinning weights. Further details are given in Sec.~\ref{subsec:graph_blocks}.

\subsection{Neighbourhood Construction} \label{subsec:graph_construction}

Identifying the neighbours of a node is an important step of Graph Convolutional Networks, as connections between nodes (edges) act as the receptive field on conventional CNNs. SkinningNet uses different strategies to create neighbourhoods, specific to each type of graph: i.e. \emph{mesh}, \emph{skeleton} and \emph{mesh-skeleton} graphs. 
In the mesh graph, faces are converted into undirected edges. Additionally, a radius-based neighbourhood is created over the mesh structure, where the k-random nodes inside of a radius $r$ are connected to the central point of the neighbourhood. The radius-based neighbourhood is used by the MUNEGC operation. In the skeleton graph, the bones are converted to undirected edges. Finally, the relation created by the \emph{Skin Binding} step is used to decide the connections in the mesh-skeleton graph.  

The \emph{Skin Binding} step is in charge of assigning which of the joints are going to influence each of the vertices. As stated in the related work section, previous works~\cite{RigNet, Liu2019} have based the skin binding process on a bone representation, where each vertex has a set of bones assigned. This assignation is done using a k-NN approach that associates the $k$ nearest bones to each of the vertices. In this work, the proposed \emph{Skin Binding} is based on a joint representation that is more natural than the bone representation. It uses the closest bones to select which are going to be the joints that influence each of the vertices. In particular, for each vertex of the mesh, the closest bones of the skeleton are found and the associated root joint for each bone is selected. Finally, that selection is refined leaving only the $k$ unique joints. The entire algorithm is described in Algorithm~\ref{alg:kjoints}.

\begin{algorithm}
\caption{k-unique joints of the closest bones}\label{alg:kjoints}
\textbf{Input:} Vertex positions, joints position and bones \\
\textbf{Output:} The selected joints for each vertex\\
\ForEach{$vertex$}{
\textbf{Compute} distance $d(vertex, bones)$\\
\textbf{Sort} distances $d$\\
\textbf{Replace} bones with their associated root joint\\
\textbf{Select} the k unique joints \\
}
\end{algorithm}

\subsection{Multi-Aggregator Graph Convolution (MAGC)} \label{subsec:magc}

\begin{figure}[t]
 \centering
 \includegraphics[width=0.45\textwidth]{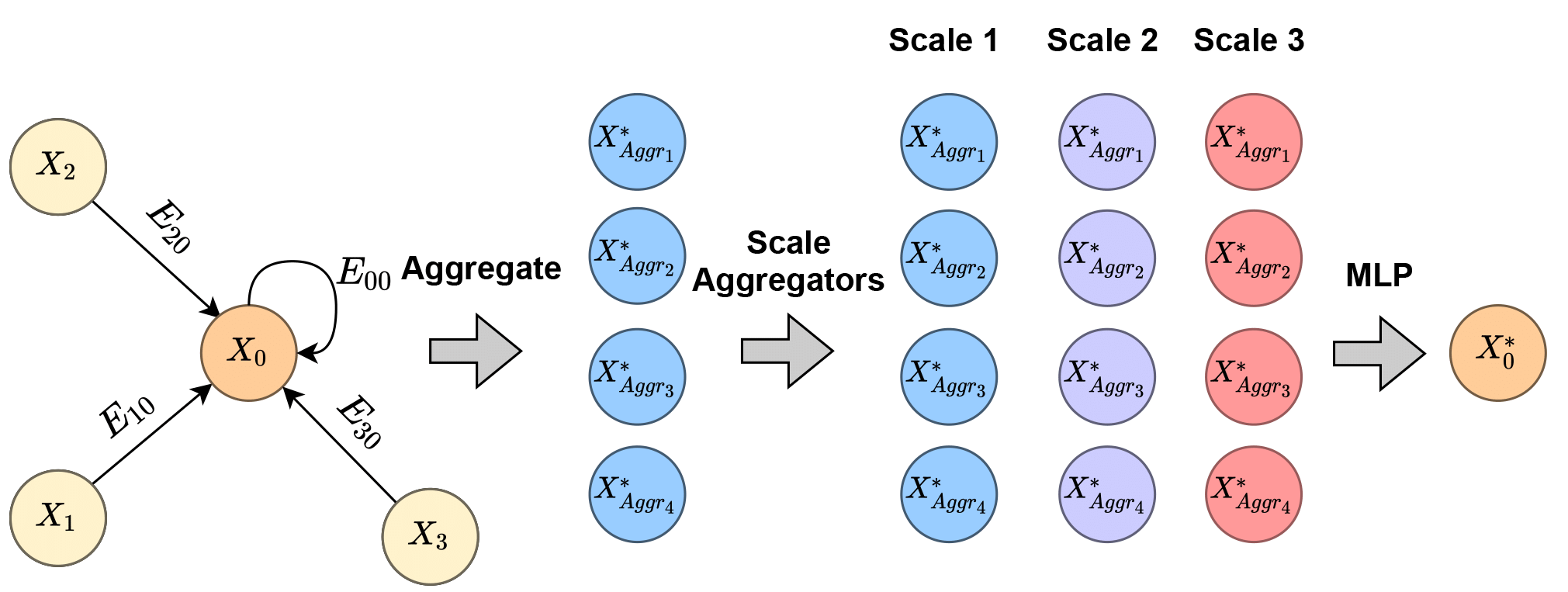}
 \caption{\textbf{Multi-Aggregator Graph Convolution} workflow. Where $E_{ji}$ is the message sent from node $j$ to node $i$.}
 \label{fig:magc}
\end{figure}

The Multi-Aggregator Graph Convolution (MAGC) is an extension of the Message-Passing scheme~\cite{messagepassing}, where multiple aggregators are used to let the graph convolution layer distinguish between neighbourhoods with identical features but with different cardinalities. The workflow of the MAGC is depicted in Fig.~\ref{fig:magc}. The first step is to compute the messages that each of the neighbours are sending to the central node of the neighbourhood. These messages are a function based on the attributes of each edge $E_{ji}=\mathcal{F}(X_j, X_i)$ where $X_j$ and $X_i$ denote the features of nodes $j$ and $i$ respectively. The messages are combined using different aggregators. The aggregators proposed in this work are formally defined in Eq.~\ref{eq:aggr} where $N(i)$ represents the neighbourhood of the node $i$. 
\begin{equation}
A=
\left\{
\begin{aligned}
A^{max} = \underset{j \in N(i)}{max}(E_{ji})\\
A^{min} = \underset{j \in N(i)}{min}(E_{ji})\\
A^{mean} = \underset{j \in N(i)}{mean}(E_{ji})\\
A^{std} = \underset{j \in N(i)}{std}(E_{ji})\\
\end{aligned}
\right.
\label{eq:aggr}
\end{equation}

The results of each aggregator are scaled using a set of logarithmic degree scalers. The proposed scalers are: \emph{i) identity}, the value of the aggregator is not changed; \emph{ii) amplification}, the value of the aggregator is amplified, and \emph{iii) attenuation}, the value of each aggregator is attenuated. Eq.~\ref{eq:scalers} formalizes the proposed scalers where $d_{train}$ refers to the mean degree of the whole training split. The motivation for the use of different logarithmic scalers is to improve the generalization of the convolution for unseen topologies, avoiding that the value of each aggregation explodes when the degree of the neighbourhood increases.
\begin{equation}
S=
\left\{
\begin{aligned}
S^{amp} = \frac{log(d)}{log(d_{train})} \\
S^{att} = \frac{log(d_{train})}{log(d)} \\
S^{iden} = 1 \\
\end{aligned}
\right.
\label{eq:scalers}
\end{equation}

Finally, the resulting operation of applying each scaler to each aggregator is fed into an MLP that learns how to fuse the information. Eq.~\ref{eq:combination} defines the combination of aggregations and scalers, with $A$ being the set of aggregation operations and $S$ the set of scaler operations. The combination of two sets of operations is described as $\otimes$ so $\mathcal{M}$ is the combination of all the scalers with all the aggregators.

\begin{equation}
\left.
\begin{aligned}
A=\{A^{max}, A^{min}, A^{mean}, A^{std}\}\\
S=\{S^{iden},S^{amp}, S^{att}\} \\
\mathcal{M} = {S \otimes A}
\end{aligned}
\right\}
\label{eq:combination}
\end{equation}

The resulting MAGC layer is described in Eq.~\ref{eq:magc}, where all combinations $\mathcal{M}$ are fused using an MLP network to produce the output feature of node $i$ and layer $l$ in a feed-forward neural network.
\begin{equation}
X^l_i = \text{MLP}\left(\underset{j \in N(i)}{\mathcal{M}}\left(E_{ji}\right)\right)
\label{eq:magc}
\end{equation}

\subsection{Graph Convolutional Blocks} \label{subsec:graph_blocks}

The proposed architecture extends the previously defined MAGC, producing three types of \emph{Graph Convolutional Blocks}: Residual MAGC, Mesh-Skeleton MAGC and Multi-Neighbourhood Graph Convolution (MUNEGC).

The \textbf{Residual MAGC} is based on the ideas proposed in~\cite{ragc, li2019deepgcns, Dang_2021_ICCV}. Each Residual MAGC is composed of two MAGCs stacked together with a short connection. In the case that the input and the output have different dimensionalities, the short connection contains a function $\mathcal{P}$ that projects the input feature space to the output ones. The Residual MAGC is formalized in Eq.~\ref{eq:rmagc} where  $\left\{\text{MAGC}_s\right\}$ represents $s$ stacked MAGC layers inside the correspondent block.
\begin{equation}
    X_i^l = \left\{\text{MAGC}_s\right\} + \mathcal{P}\left(X_i^{l-1}\right)
    \label{eq:rmagc}
\end{equation}

The \textbf{Mesh-Skeleton Graph Convolution} employs a modification of MAGC to deal with graphs where the neighbourhood is composed of an heterogeneous combination of nodes. To combine the mesh and the skeleton nodes, a one-hot vector is concatenated to the node feature. In this way, features corresponding to the mesh node can be distinguished from features corresponding to skeleton nodes by the graph operation. Eq.~\ref{eq:hmagc} shows the one-hot concatenation proposed for the combination:
\begin{equation}
X_i^l = \left\{
\begin{tabular}{l}
$\text{concat}\left\{[0], X_i^l\right\} \quad \text{if} \quad i \in \text{mesh}$ \\
$\text{concat}\left\{[1] ,X_i^l\right\} \quad \text{if} \quad i \in \text{skeleton}$ \\
\end{tabular}
\right.
\label{eq:hmagc}
\end{equation}

The \textbf{Multi-Neighbourhood Graph Convolution} is an extension of MAGC based on~\cite{MOSELLAMONTORO2021} where two kinds of neighbourhoods are used to get the new node feature. Eq.~\ref{eq:munegc_magc} defines the extension of MAGC to a multi-neighbourhood approach. Where $\mathcal{K}$ is the set of neighbourhood types:
\begin{equation}
\begin{split}
X_i^l = \text{MLP}\left( \underset{k \in \mathcal{K}}{\text{concat}}\left\{ \text{MLP}\left(\underset{(j) \in N^k(i)}{\mathcal{M}}\left(E^{k}_{ji}\right)\right) \right\} \right)
\end{split}
\label{eq:munegc_magc}
\end{equation}
\section{Results}
\label{sec:results}
\subsection{Dataset} \label{subsec:dataset}

The proposed method has been trained and evaluated on the \textbf{RigNetv1}~\cite{RigNet} dataset which is publicly accessible for non-commercial use. This dataset is composed of $2703$ rigged characters of different categories. The original split of the dataset is followed where $2163$ assets are used for training, $270$ for validation and $270$ for testing. All training assets contain between $1$k and $5$k vertices, and all of them are scaled between $[-1,1]$ and oriented to face the same direction. The number of skeleton joints is in a range from $3$ to $48$ with a mean of $25$. In addition, two more assets from the \textbf{Paragon Collection}~\cite{paragondata} have been used for the generalization study in Sec.~\ref{subsec:genstudy}. These two assets have been simplified so the resulting meshes contain about 6k vertices and 50 joints and normalized as done in \textbf{RigNetv1}. The \textbf{Paragon Collection} dataset is allowed to be used in non-interactive linear media products under a non-exclusive and non-transferable license.

\subsection{Implementation Details} \label{subsec:implementation_details}
The detailed architecture with the number of filters used in each layer is shown in Table~\ref{tab:arch_details}. Several attributes are selected as initial features for nodes in the mesh and skeleton structures to help the network to learn the relation between both structures. The node attributes used to describe the mesh are the normalized 3D coordinates of the vertices, the Geodesic distance to the $k=5$ unique joints from the nearest bones, the start and end positions of the bone of those joints and a boolean value indicating if the joint is an end joint or not. In the case of the skeleton, only the normalized 3D position of the joints is included. Sec.~\ref{subsec:eucvsgeo} discusses the advantages and disadvantages of using the Geodesic distance with respect to the Euclidean. To compute the Geodesic distance between joints and vertices, the volumetric version proposed in~\cite{RigNet} is employed, where the shortest path from vertex to joint passing through the interior mesh volume is computed.

The radius for the second neighbourhood used in the Skinning Prediction Network is $r=0.06$. From all nodes inside the radius $r$, a maximum of $10$ nodes are randomly sampled to create the neighbourhood. The final output of SkinningNet is the skinning weights for the $k=5$ unique joints of the closest bones of each vertex. A dropout layer is added before each Fully Connected layer of the Skinning Prediction Network with a probability of $p=0.5$. The network is trained in an end-to-end fashion for $200$ epochs. The Rectified Adam (RAdam)~\cite{radam} optimizer is used with a learning rate of $1\times10^{-4}$, a weight decay of $1\times10^{-4}$, and a batch size of $4$. The Kullback-Leibler divergence loss is used to minimize the distance between the predicted skinning weight distribution and the ground truth distribution.   

\begin{table}[ht]
\begin{center}
{\footnotesize
\begin{tabular}{|c|c|}
\hline 
\multicolumn{2}{|c|}{\textbf{Mesh Network}} \\
\hline 
\textbf{Layer} & \textbf{N.Filters}\\ 
\hline 
Input Transform & MLP(64, 128) \\
Residual MAGC & 128 \\
Residual MAGC & 256 \\
Residual MAGC & 512 \\
\hline 
\multicolumn{2}{|c|}{\textbf{Skeleton Network}} \\
\hline 
Input Transform & MLP(64) \\
MAGC & 128 \\
MAGC & 256 \\
MAGC & 512 \\
\hline
\multicolumn{2}{|c|}{\textbf{Mesh - Skeleton Network}} \\
\hline 
Mesh Global Shape & MLP(256) \\
Skeleton Global Shape & MLP(256) \\
Mesh-Skel MAGC & 512 \\
Concat & 512 + 256 +  256 \\
\hline 
\multicolumn{2}{|c|}{\textbf{Skinning Network}} \\
\hline 
MAGC & 256 \\
MAGC & 128 \\
MAGC & 64 \\
MLP & (64, 32, k) \\
\hline 
\end{tabular}
}
\end{center}
\caption{\textbf{SkinningNet} architecture details with the number of filters used in each layer. k is the number of joints that can influence each of the vertices.}
\label{tab:arch_details}
\end{table}

\subsection{Comparison With The State-of-the-art}

SkinningNet is compared to the two most recent data-driven approaches: NeuroSkinning~\cite{Liu2019} and RigNet~\cite{RigNet}. Both networks have been trained from scratch following the procedure described in their respective papers. In the case of NeuroSkinning, the original Euclidean distance has been replaced with the Geodesic one, as it will be demonstrated in Sec.~\ref{subsec:eucvsgeo} to be a better choice for watertight meshes. Furthermore, to guarantee a fair comparison, the same input attributes used in SkinningNet and described in Sec.~\ref{subsec:implementation_details} have been used in the two architectures.
Both methods proposed a deep learning architecture based on graph convolutional layers. In the case of NeuroSkinning the Graph Attention layer~\cite{GAT2018} is used, whereas RigNet uses the Edge Convolution layer~\cite{dgcnn}.
Four different metrics have been used to evaluate the skinning weight prediction: \\
\noindent1. \emph{Precision and Recall} finds the set of joints that influence each vertex significantly, where influence corresponds to a prediction larger than $1\times10^{-4}$, as described in~\cite{RigNet, Liu2019}.\\
\noindent2. \emph{L1-norm} of the difference between the predicted skinning weight and the ground truth vectors of each vertex of the mesh. The average of this metric is computed across the whole test split.\\
\noindent3. \emph{Deformation error} computes the Euclidean distance between the position of the vertices deformed after applying the predicted skinning weights and the ground truth ones. To compute this metric, $10$ different random poses are computed where all the joints in the skeleton are randomly rotated within a range of $\pm10$ degrees.

\begin{figure}[ht]
 \centering
 {\footnotesize
\begin{tabular}{c}
 \includegraphics[width=0.45\textwidth]{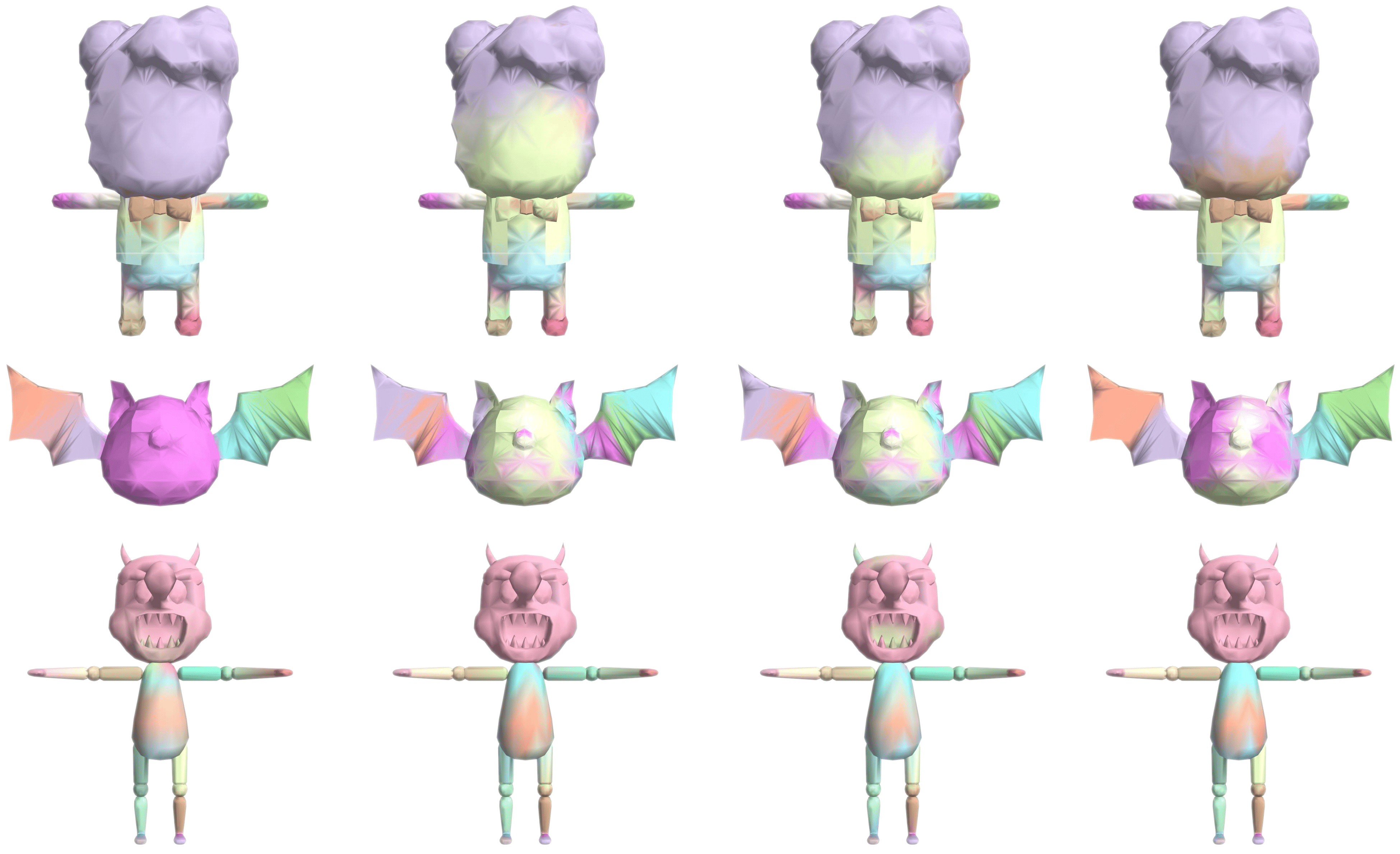} \\
 \hspace{2mm} (a) GT \hspace{4mm} (b) NeuroSkinning \hspace{3mm} (c) RigNet \hspace{4mm} (d) SkinningNet \\
\end{tabular}
}
 \caption{\textbf{Skinning weights} prediction results of each of the state-of-the-art methods. A random color is assigned to each joint and colors are blended between vertices in the mesh using the skinning weights.}
 \label{fig:skinning_rignetdata}
\end{figure}

\emph{SkinningNet} outperforms the best method of the state-of-the-art with over $5\%$ of improvement on Precision with the same Recall and $15\%$ improvement on average L1-norm. Table~\ref{tab:sota-results-predictions} summarizes the comparison with the state-of-the-art. Fig.~\ref{fig:skinning_rignetdata} shows the skinning weight prediction examples on three assets. A random color is assigned to each skeleton joint and colors are blended between vertices in the mesh using the skinning weights. It can be seen, for instance in the \emph{bat} asset in the middle row, how the proposed algorithm better predicts skinning weights with associated colors that are closer to those of the ground truth in the first column.

\begin{table}[ht]
\begin{center}
\begin{tabular}{|c|c|c|c|}
\hline 
\textbf{Method} & \textbf{Prec(\%)} & \textbf{Rec.(\%)} & \textbf{avg L1}\\ 
\hline \hline
NeuroSkinning~\cite{Liu2019} & 82.3 & 79.7 & 0.41  \\
Rignet~\cite{RigNet} & 82.3 & \textbf{80.8} & 0.39  \\
\textbf{SkinningNet} & \textbf{87.0} & \textbf{80.8} & \textbf{0.33}  \\
\hline
\end{tabular}
\end{center}
\caption{\textbf{Prediction results} comparison with the current state-of-the-art techniques.}
\label{tab:sota-results-predictions}
\end{table}

\begin{figure}[ht]
 \centering
 {\footnotesize
\begin{tabular}{l}
 \includegraphics[width=0.45\textwidth]{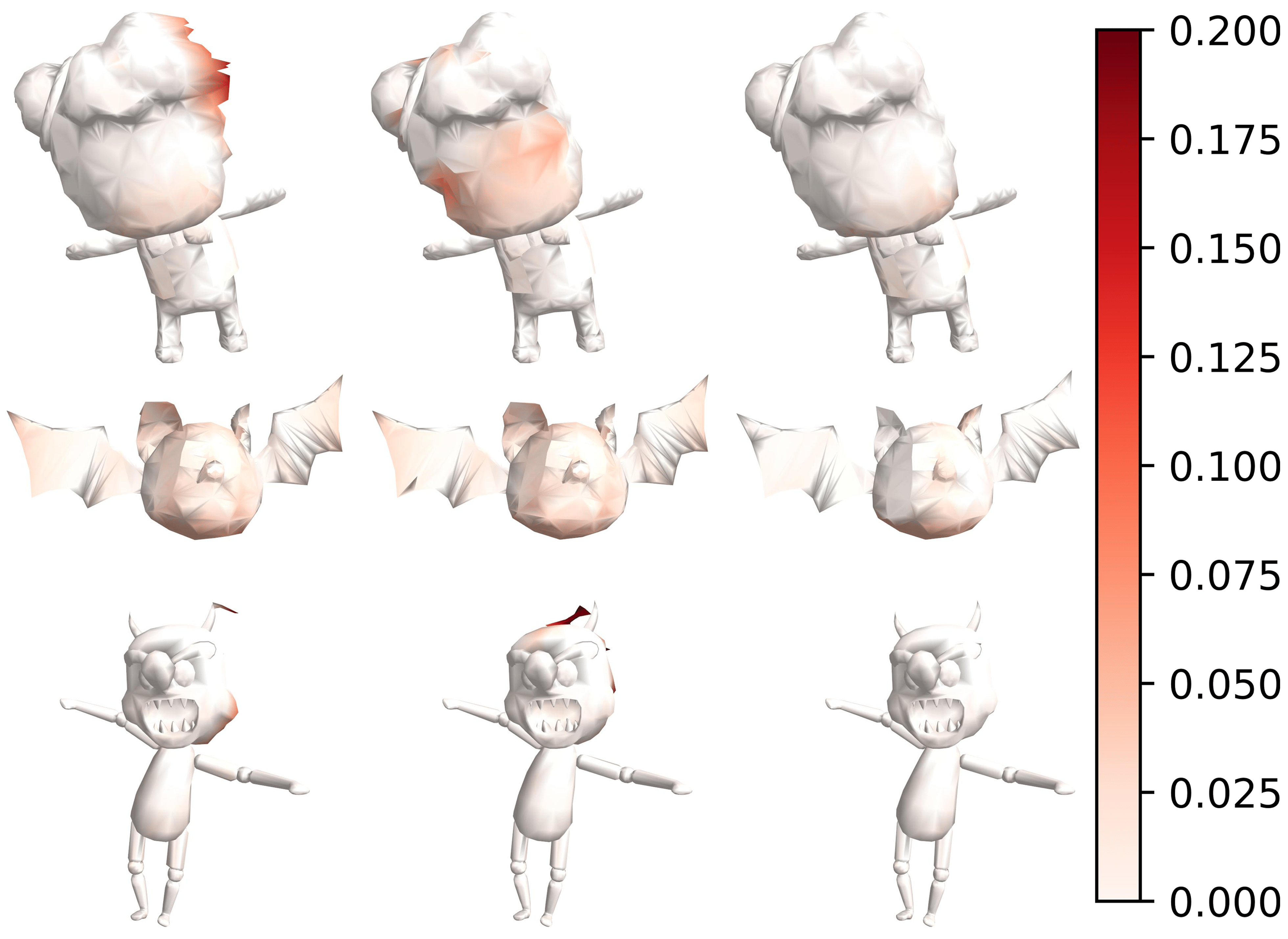} \\
 (a) NeuroSkinning \hspace{4mm} (b) RigNet \hspace{8mm} (c) SkinningNet \\
\end{tabular}
}
\caption{\textbf{Deformation error} of three different characters with a randomly generated pose.}
\label{fig:weightmap_rignetdata}
\end{figure}

Even though results are significantly better than state-of-the-art, the Precision, Recall and L1-norm metrics are not good enough to evaluate skinning weight predictions. The reason is that, in the case of Precision and Recall, the magnitude of the skinning weight is not taken into account. Furthermore, in the case of the L1-norm, it does not completely capture the importance of an error on the skinning weights. Two similar L1 errors can result on different deformation errors depending on whether the L1 error comes from small differences on each of the associated joints or it is concentrated on a single joint. For these reasons, the deformation error will be used for the rest of the experiments. Table~\ref{tab:sota-results-deformation} shows the average and the maximum deformation error. Again, SkinningNet outperforms with a $20\%$ of improvement in the average error and with a $17\%$ of improvement in the maximum error. The qualitative results of this metric can be observed in Fig.~\ref{fig:weightmap_rignetdata}, where SkinningNet is able to generate reasonable results where previous state-of-the-art methods fail. 

\begin{table}[ht]
\begin{center}
\begin{tabular}{|c|c|c|}
\hline 
\textbf{Method} & \textbf{Avg. Def} & \textbf{Max. Def}\\ 
\hline \hline
NeuroSkinning~\cite{Liu2019} & 0.002843 & 0.2151 \\
Rignet~\cite{RigNet} & 0.002921 & 0.2246  \\
\textbf{SkinningNet} & \textbf{0.002288} & \textbf{0.1789} \\
\hline
\end{tabular}
\end{center}
\caption{\textbf{Deformation error} comparison with the current state-of-the-art techniques. The errors are inside a range of $[0,0.2]$ and are calculated over the normalized version of each mesh.}
\label{tab:sota-results-deformation}
\end{table}

\subsection{Architecture Design Study}
In this section the influence of each of the components of the network is studied. Table~\ref{tab:ablation-arch} shows the results of removing the global shape feature, the residual connections and the MUNEGC from the original SkinningNet. 

\begin{table}[ht]
\begin{center}
\begin{tabular}{|c|c|c|}
\hline 
\textbf{Method} & \textbf{Avg. Def} & \textbf{Max. Def}\\ 
\hline \hline
\textbf{Baseline} & \textbf{0.002288} & \textbf{0.1789}  \\
No Global Feature & 0.002452 & 0.1905 \\
No Residual  & 0.002394 & 0.1862 \\
No MUNEGC & 0.002427 & 0.2009 \\
\hline
\end{tabular}
\end{center}
\caption{\textbf{Architecture Study}  that shows the influence of each of the proposed stages in the output.}
\label{tab:ablation-arch}
\end{table}

Removing the MUNEGC layers from the network leads to a loss of performance of about $5\%$. These results show that MUNEGC is helping to have an enriched local descriptor for each of the vertices. This descriptor enables the network to be aware of the local structure around the vertices. 
Similar performance losses are obtained when removing the \emph{Global Shape Feature} as it helps the network to be aware of the global shape of the skeleton and the mesh when doing the prediction. Finally, adding the residual connections to the \emph{Mesh Network} leads to an improvement of $4\%$, demonstrating that the residual connections are helping the proposed approach.

\subsection{Joint vs.\ Bone Representation Study}

In this section, the difference between a joint vs.\ a bone representation is analyzed. The \emph{Skin Binding} step is modified to use bones instead of joints when creating the relations between the mesh and the skeleton. Table~\ref{tab:ablation-joint-bone} shows the results of both approaches. As can be seen, the skeleton joint representation gives an improvement of $5\%$ in both average and maximum deformation. This improvement is due to the fact that the joint representation follows a natural approach where each of the joints represents an articulation which is in charge of defining the movement of each bone. 

\begin{table}[ht]
\begin{center}
\begin{tabular}{|c|c|c|}
\hline 
\textbf{Method} & \textbf{Avg. Def} & \textbf{Max. Def}\\ 
\hline \hline
\textbf{Joint} & \textbf{0.002288} & \textbf{0.1789}  \\
Bone & 0.002407 & 0.1893 \\
\hline
\end{tabular}
\end{center}
\caption{\textbf{Joint vs.\ Bone} representation study, where the influence of each representation is analyzed.}
\label{tab:ablation-joint-bone}
\end{table}

\subsection{Euclidean vs.\ Geodesic Distance Study}\label{subsec:eucvsgeo}

Finding the vertex to joint distance is a critical step for skinning prediction methods. NeuroSkinning~\cite{Liu2019} proposed the use of the Euclidean distance while RigNet~\cite{RigNet} proposed to use the Geodesic distance. Both distances have their advantages and disadvantages. The Geodesic distance is defined for connected components, with the distance between two non-connected components being infinity. This means that the Geodesic distance is better suited for watertight meshes, whereas the Euclidean distance can be used for both watertight and non-watertight meshes. In terms of performance, using Geodesic distance helps the network to predict better results than using Euclidean distance as observed in Table~\ref{tab:ablation-distances}.  

\begin{table}[ht]
\begin{center}
\begin{tabular}{|c|c|c|}
\hline 
\textbf{Method} & \textbf{Avg. Def} & \textbf{Max. Def}\\ 
\hline \hline
Euclidean & 0.002663 & 0.4473 \\
\textbf{Geodesic} & \textbf{0.002288} & \textbf{0.1789}  \\
\hline
\end{tabular}
\end{center}
\caption{\textbf{Geodesic vs.\ Euclidean} performance comparison. }
\label{tab:ablation-distances}
\end{table}

\begin{figure}[t]
\begin{center}
\begin{tabular}{ccc}
\includegraphics[width=0.12\textwidth]{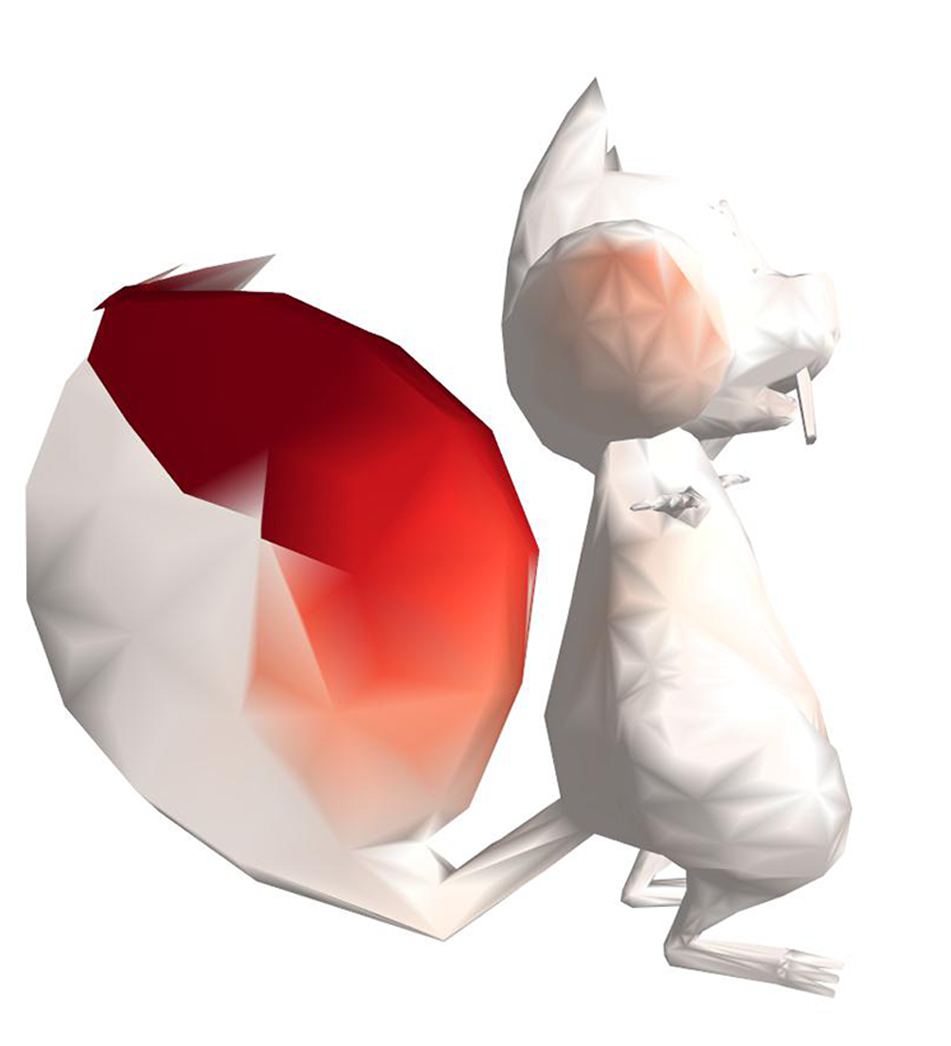} & 
         \includegraphics[width=0.12\textwidth]{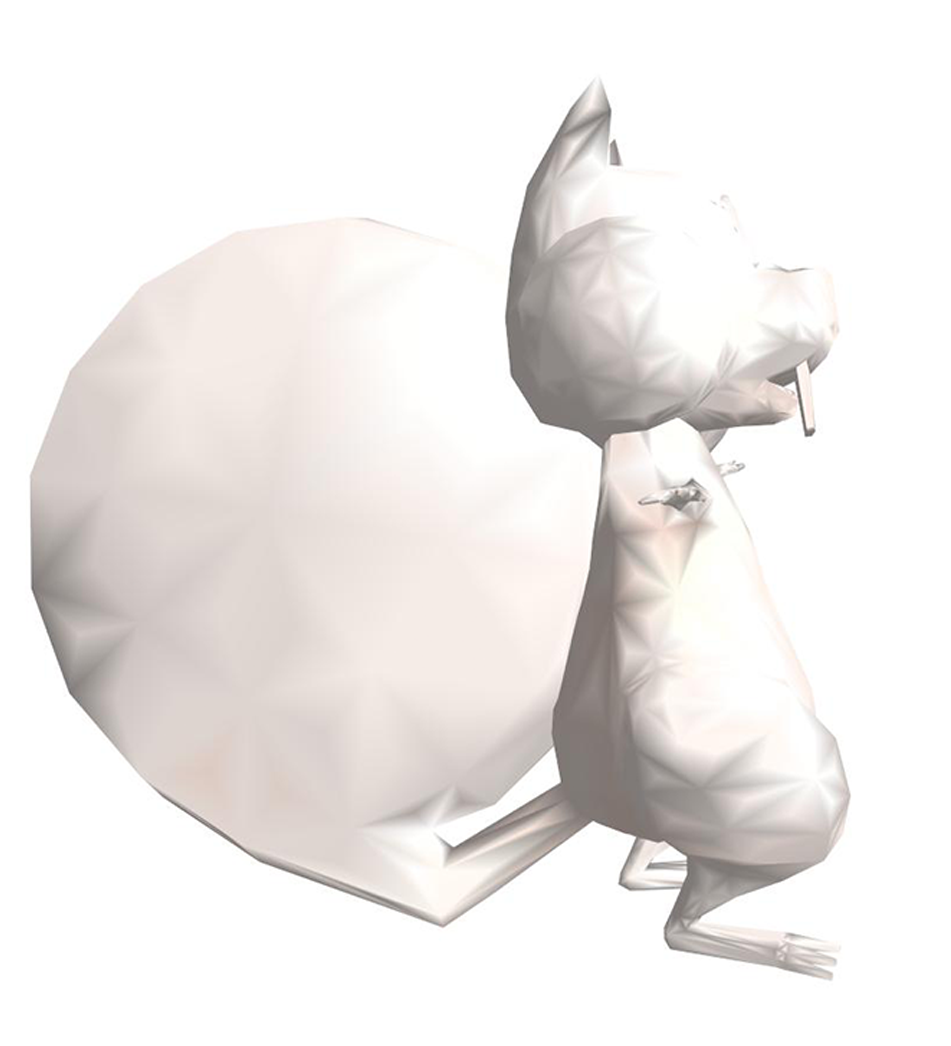} & 
         \includegraphics[width=0.035\textwidth]{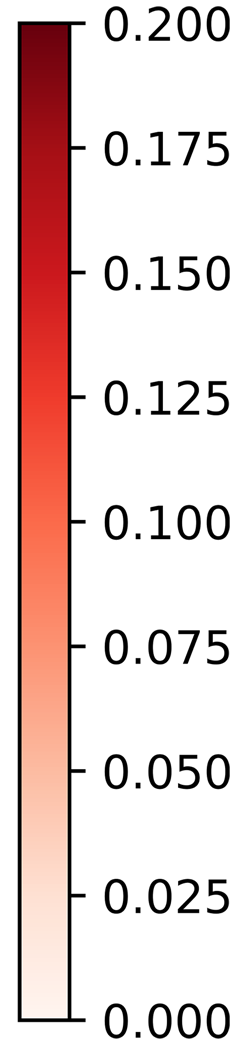} \\
{\footnotesize (a)  Euclidean} & {\footnotesize (b) Geodesic } &
\end{tabular}
\caption{\textbf{Euclidean vs.\ Geodesic} qualitative results. It can be observed that the euclidean distance introduces big deformation errors in the tail.}
\label{fig:euclideanvsgeo}
\end{center}
\end{figure}

An example of the effects of a higher maximum error when using the Euclidean distance can be observed in Fig.~\ref{fig:euclideanvsgeo} where the tail of the squirrel is deformed with respect to the Geodesic result.

\subsection{Graph Convolution Study}\label{subsec:graphconvstudy}
In this work, the use of the Multi-Aggregator Graph Convolution (MAGC) is proposed. To study the influence of this operator on the output of the prediction, the MAGC has been replaced by three different state-of-the-art graph convolutions. Table~\ref{tab:graphstudy} details the results of this study, showing that Edge Convolution~\cite{dgcnn} achieves the closest results to the proposed MAGC. The improvement in respect to the Edge Convolution is about $4\%$ on the average deformation and about $6\%$ on the max deformation, demonstrating that the MAGC helps to predict better skinning weights.

\begin{table}[ht]
\begin{center}
\begin{tabular}{|c|c|c|}
\hline 
\textbf{Method} & \textbf{Avg. Def} & \textbf{Max. Def}\\ 
\hline \hline
\textbf{MAGC} & \textbf{0.002288} & \textbf{0.1789} \\
EdgeConv~\cite{dgcnn} & 0.002381 & 0.1921 \\
GAT~\cite{GAT2018} & 0.002551 & 0.2098 \\
GCN~\cite{kipf2017semi} & 0.002765 & 0.2533 \\
\hline
\end{tabular}
\end{center}
\caption{\textbf{Graph Convolution} study where MAGC has been compared with three different state-of-the-art operators.}
\label{tab:graphstudy}
\end{table}

\subsection{Generalization Study}\label{subsec:genstudy}
A generalization study was performed to evaluate if the proposed method has good generalization and is suitable for working with AAA game characters. The trained networks using RigNetv1~\cite{RigNet} have been applied to two high quality assets from the Paragon Collection~\cite{paragondata}. 

\begin{table}[ht]
\begin{center}
\begin{tabular}{|c|c|c|}
\hline 
\textbf{Method} & \textbf{Avg. Def} & \textbf{Max. Def}\\ 
\hline \hline
NeuroSkinning & 0.003724 & 0.1213 \\
Rignet & 0.003398 & 0.1051 \\
\textbf{SkinningNet} & \textbf{0.002666} &\textbf{ 0.0664} \\
\hline
\end{tabular}
\end{center}
\caption{\textbf{Generalization Study} of the state-of-the-art methods. The deformation error is computed using a normalized version of the Paragon Assets~\cite{paragondata}.}
\label{tab:generalization_study}
\end{table}

\begin{figure}[ht]
 \centering
 {\footnotesize
\begin{tabular}{l}
 \includegraphics[width=0.45\textwidth]{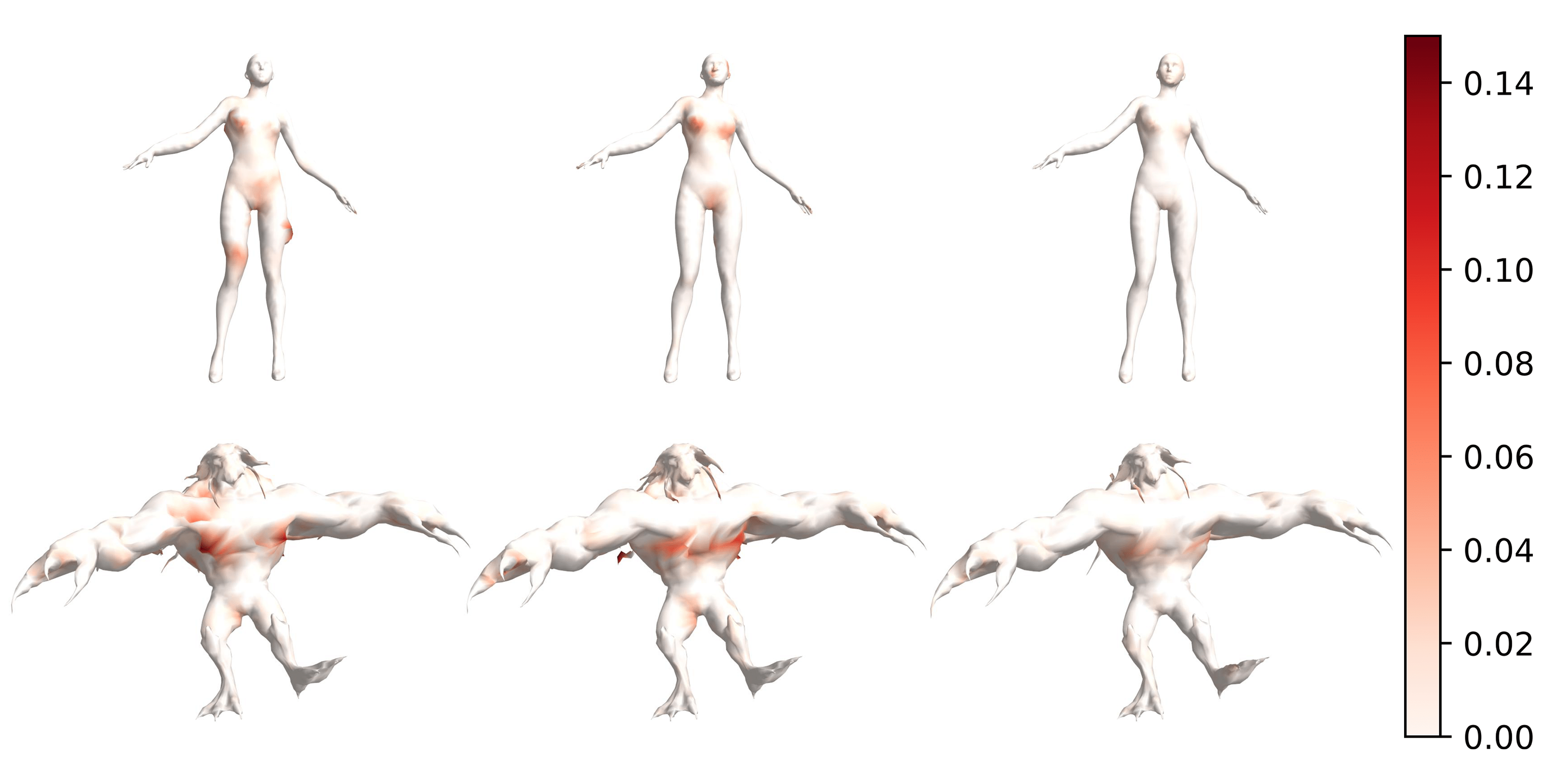} \\
 {\footnotesize(a) NeuroSkinning \hspace{8mm} (b) RigNet \hspace{7mm} (c) SkinningNet }\\
\end{tabular}
}
\caption{\textbf{Generalization study} with Aurora and Rampage assets from Paragon collection~\cite{paragondata}.}
\label{fig:weightmap_paragon}
\end{figure}

Results in Table~\ref{tab:generalization_study} show that SkinningNet outperforms previous works with over $28\%$ improvement in average deformation and $36\%$ in maximum deformation error, proving that the proposed method generalizes better than previous methods for unseen complex characters. In Fig.~\ref{fig:weightmap_paragon}, it can be observed that the proposed method is generating good quality animations without strong errors, whereas the other methods have high errors in both characters. 

\section{Conclusions and Limitations}
\label{sec:conclusions}

This work presents SkinningNet, a Two-Stream Graph Convolutional Neural Network that automatically generates skinning weights for an input mesh and its associated skeleton. The SkinningNet architecture is based on the novel Multi-Aggregator Graph Convolution layer that allows the network to better generalize for unseen topologies. Moreover, the proposed joint-based skin binding and Mesh-Skeleton network learns to find the best relationships between the mesh and skeleton helping to improve the final skinning weight predictions. The proposed architecture outperforms current approaches with over a $20\%$ improvement on mesh deformation error and is also able to better generalize for complex characters of unseen domains.

Even though results are promising there are some limitations that could be addressed. The first one is that assigning the joints that will influence each of the vertices is based on a k-NN approach. The network can learn which of these joints will affect each vertex, however, if the joint that should influence the vertex is not inside the initial k proposal, the network will not be able to find it. To overcome this issue, we propose to explore link prediction strategies to let the network learn the binding strategy. The second one is that the current configuration uses the standard Geodesic distance that, by definition, is infinity on non-connected regions, meaning that the network could experience a drop on its performance in non-watertight meshes. To overcome this limitation, new approximations to the Geodesic distance using voxelization could be explored. Finally, the use of the deformation error as a loss could be explored to improve the performance of the current approach.

{\footnotesize
    \textbf{Acknowledgements.}
    This work has been partially supported by the project PID2020-117142GB-I00, funded by MCIN/AEI /10.13039/501100011033. The authors would like to thank Denis Tome for his technical advice during the development of this project.
    
}

{\small
\bibliographystyle{ieee_fullname}
\bibliography{egbib}
}

\end{document}